\documentclass[10pt,twocolumn,letterpaper]{article}

\usepackage{cvpr}
\usepackage{times}
\usepackage{epsfig}
\usepackage{graphicx}
\usepackage{amsmath}
\usepackage{amssymb}
\usepackage{pifont}
\newcommand{\cmark}{\ding{51}}%
\newcommand{\xmark}{\ding{55}}%
\usepackage{mathtools}
\DeclarePairedDelimiter\floor{\lfloor}{\rfloor}
\usepackage{bm}
\usepackage{bbm}
\usepackage{multirow}
\usepackage{multicol}
\usepackage{gensymb}
\usepackage{graphics}
\DeclareMathOperator{\arctantwo}{arctan2}
\usepackage[dvipsnames]{xcolor}
\definecolor{mygray}{gray}{15}

\usepackage[pagebackref=true,breaklinks=true,letterpaper=true,colorlinks,bookmarks=false]{hyperref}

\cvprfinalcopy 

\def\cvprPaperID{16} 

\ifcvprfinal\fi
\begin{document}

\title{AFDet: Anchor Free One Stage 3D Object Detection}

\author{
Runzhou Ge$^{*}$ \quad Zhuangzhuang Ding$^{*}$ \quad Yihan Hu$^{*}$ \\
Yu Wang \quad Sijia Chen \quad Li Huang \quad Yuan Li \\
Horizon Robotics\\
{\tt\small \{runzhou.ge, zhuangzhuang.ding, yihan01.hu, yu04.wang\}@horizon.ai} 
}

\maketitle

\newcommand\blfootnote[1]{%
  \begingroup
  \renewcommand\thefootnote{}\footnote{#1}%
  \addtocounter{footnote}{-1}%
  \endgroup
}

\blfootnote{$\prescript{*}{}{}$These authors contributed equally to this work.}

\begin{abstract}

High-efficiency point cloud 3D object detection operated on embedded systems is important for many robotics applications including autonomous driving. Most previous works try to solve it using anchor-based detection methods which come with two drawbacks: post-processing is relatively complex and computationally expensive; tuning anchor parameters is tricky. We are the first to address these drawbacks with an anchor free and Non-Maximum Suppression free one stage detector called AFDet. The entire AFDet can be processed efficiently on a CNN accelerator or a GPU with the simplified post-processing. Without bells and whistles, our proposed AFDet performs competitively with other one stage anchor-based methods on KITTI validation set and Waymo Open Dataset validation set.
\end{abstract}


\section{Introduction}

Detecting 3D objects in the point cloud is one of the most important perception tasks for autonomous driving.
To satisfy the power and efficiency constraints, most of the detection systems are operated on vehicle embedded systems. Developing embedded systems friendly point cloud 3D detection system is a critical step to make autonomous driving a reality.

Due to the sparse nature of the point cloud, it is inefficient to directly apply 3D or 2D Convolution Neural Networks (CNN)~\cite{krizhevsky2012imagenet, simonyan2014very} on the raw point cloud. On one hand, lots of point cloud encoders~\cite{zhou2018voxelnet, chen2017multi, yan2018second, lang2019pointpillars, zhou2019end} are introduced to encode the raw point cloud to data formats that could be efficiently processed by 3D or 2D CNN. On the other hand, some work~\cite{qi2018frustum, wang2019frustum, shi2019pointrcnn, yang2019std, ngiam2019starnet} directly extract features from raw point clouds for 3D detection which is inspired by PointNet~\cite{qi2017pointnet, qi2017pointnet++}. But for the detection part, most of them adopt anchor-based detection methods proven effective in image object detection tasks.

\begin{table}[t]
\begin{center}
\setlength\tabcolsep{4pt}
\resizebox{\columnwidth}{!}{%
\begin{tabular}{l|cc}
\hline
                           &Anchor-based               &AFDet (Ours)   \\
\hline\hline
Anchor Free                &\xmark                             &\cmark    \\
NMS Free                     &\xmark                           &\cmark \\
Post-processing Friendly       &\xmark                             &\cmark    \\
Embedded Systems Friendly               &\xmark                             &\cmark    \\
\hline
\end{tabular}
}
\end{center}
\caption{The comparison between anchor-based methods and our method. We use max pooling and AND operation to achieve a similar functionality with NMS but with a much higher speed. In our experiments, our max pooling and AND operation can achieve $2.5 \times 10^{-5}$ s on one Nvidia 2080 Ti GPU which is approximately $1000\times$ faster than the CPU implemented NMS.} 
\label{tbl:comparison}
\end{table}

Anchor-based methods have two major disadvantages. First, Non-Maximum Suppression (NMS) is necessary for anchor-based methods to suppress the overlapped high confident detection bounding boxes. But it can introduce non-trivial computational cost especially for embedded systems. According to our experiments, it takes more than 20 ms to process one KITTI~\cite{Geiger2012CVPR} point cloud frame even on a modern high-end desktop CPU with an efficient implementation, let alone CPUs typically deployed for embedded systems. Second, anchor-based methods requires anchor selection which is tricky and time-consuming, because critical parts of the tuning can be a manual trial and error process. For instance, every time a new detection class is added to the detection system, hyper parameters such as appropriate anchor number, anchor size, anchor angle and anchor density need to be selected.

Can we get rid of NMS and design an embedded system friendly anchor free point cloud 3D detection system with high efficiency? Recently, anchor free methods~\cite{Law_2018_CornerNet, zhou2019objects, tian2019fcos} in image detection have achieved remarkable performance. In this work, we propose an anchor free and NMS free one stage end-to-end point cloud 3D object detector (AFDet) with simple post-processing.

We use PointPillars~\cite{lang2019pointpillars} to encode the entire point cloud into pseudo images or image-like feature maps in Bird's Eye View (BEV) in our experiments. However, AFDet can be used with any point cloud encoder which generates pseudo images or image-like 2D data. After encoding, a CNN with upsampling necks is applied to output the feature maps, which connect to five different heads to predict object centers in the BEV plane and to regress different attributes of the 3D bounding boxes. Finally, the outputs of the five heads are combined together to generate the detection results. A keypoint heat map prediction head is used to predict the object centers in the BEV plane. It will encode every object into a small area with a heat peak as its center. At the inference stage, every heat peak will be picked out by max pooling operation. After this, we no longer have multiple regressed anchors tiled into one location, therefore there is no need to use traditional NMS. This makes the entire detector runnable on a typical CNN accelerator or GPU, saving CPU resources for other critical tasks in autonomous driving. 

Our contributions can be summarized as below:

(1) We are the first to propose an anchor free and NMS free detector for point cloud 3D object detection with simplified post-processing.

(2) AFDet is embedded system friendly and can achieve high processing speed with much less engineering effort. 

(3) AFDet can achieve competitive accuracy compared with previous single-stage detectors on the KITTI validation set. A variant of our AFDet surpasses the state-of-the-art single-stage 3D detection methods on Waymo validation set.

In the following, we first discuss related work in Section~\ref{related_work}. Then we show more details of our method in Section~\ref{method}. Finally, we analyze and compare AFDet with other approaches in Section~\ref{experiment}.

\section{Related Work}
\label{related_work}
Thanks to accurate 3D spatial information provided by LiDAR, LiDAR-based solutions prevail in 3D object detection task.

\subsection{LiDAR-based 3D Object Detection}
Due to non-fixed length and order, point clouds are in a sparse and irregular format which needs to be encoded before input into a neural network. Some works utilize mesh grid to voxelize point clouds. Features, such as density, intensity, height \etc, are concatenated in different voxels as different channels. Voxelized point clouds are either projected to different views such as BEV, Range View (RV) \etc, to be processed by 2D convolution~\cite{chen2017multi, ku2018joint, simony2018complex, yang2018pixor} or kept in 3D coordinates to be processed by sparse 3D Convolution~\cite{song2016deep}. PointNet~\cite{qi2017pointnet} proposes an effective solution to use raw point cloud as input to conduct 3D detection and segmentation. PointNet wields Multilayer Perceptron (MLP) and max pooling operation to solve point cloud's disorder and non-uniformity and provides satisfactory performance. Successive 3D detection solutions based on the raw point cloud input provide promising performance such as PointNet++~\cite{qi2017pointnet++}, Frustum PointNet~\cite{qi2018frustum}, PointRCNN~\cite{li2018pointcnn} and STD~\cite{yang2019std}. VoxelNet~\cite{zhou2018voxelnet} combines voxelization and PointNet to propose Voxel Feature Extractor (VFE) in which a PointNet style encoder is implemented inside each voxel. A similar idea is used in SECOND~\cite{yan2018second} despite that sparse 3D convolution is utilized to further extract and downsample information in $z$-axis following VFE. VFE improves the performance of the LiDAR-based detector dramatically, however, with encoders that are learned from data, the detection pipeline becomes slower. PointPillars~\cite{lang2019pointpillars} proposes to encode point cloud as pillars instead of voxels. As a result, the whole point cloud becomes a BEV pseudo image whose channels are equivalent to VFE's output channels instead of 3.

\textbf{Anchor free.} In anchor-based methods, pre-defined boxes are provided for bounding box encoding. However, using dense anchors lead to exhaustive numbers of potential target objects, which makes NMS an unavoidable issue. Some previous work~\cite{yang2018pixor, meyer2019lasernet, chen2019object, shi2019pointrcnn, qi2019deep} mention anchor free concepts. PointRCNN~\cite{shi2019pointrcnn} proposes a 3D proposal generation sub-network without anchor boxes based on whole-scene point cloud segmentation. VoteNet~\cite{qi2019deep} constructs 3D bounding boxes from voted interest points instead of predefined anchor boxes. But all of them are not NMS free, which makes them less efficient and is not friendly to the embedded systems. Besides, PIXOR~\cite{yang2018pixor} is a BEV detector rather than a 3D detector.

\subsection{Camera-based 3D Object Detection}

Camera-based solutions thrived in accordance with the willingness of reducing cost. With more sophisticated networks being designed, camera-based solutions are catching up rapidly with LiDAR-based solutions.
MonoDIS~\cite{simonelli2019disentangling} leverages a novel disentangling transformation for 2D and 3D detection losses and a novel self-supervised confidence score for 3D bounding boxes. It gets top ranking on nuScenes~\cite{nuscenes2019} 3D object detection challenge. CenterNet~\cite{zhou2019objects} predicts the location and class of an object from the center of its bounding box on a feature map. Though originally designed for 2D detection, CenterNet also has the potential to conduct 3D detection with a mono camera. TTFNet~\cite{liu2019training} proposes techniques to shorten training time and increase inference speed. RTM3D~\cite{li2020rtm3d} predicts nine perspective keypoints of a 3D bounding box in image space and recover the 3D bounding box with geometric regulation.

\begin{figure*}
\begin{center}

\includegraphics[width=0.9\textwidth]{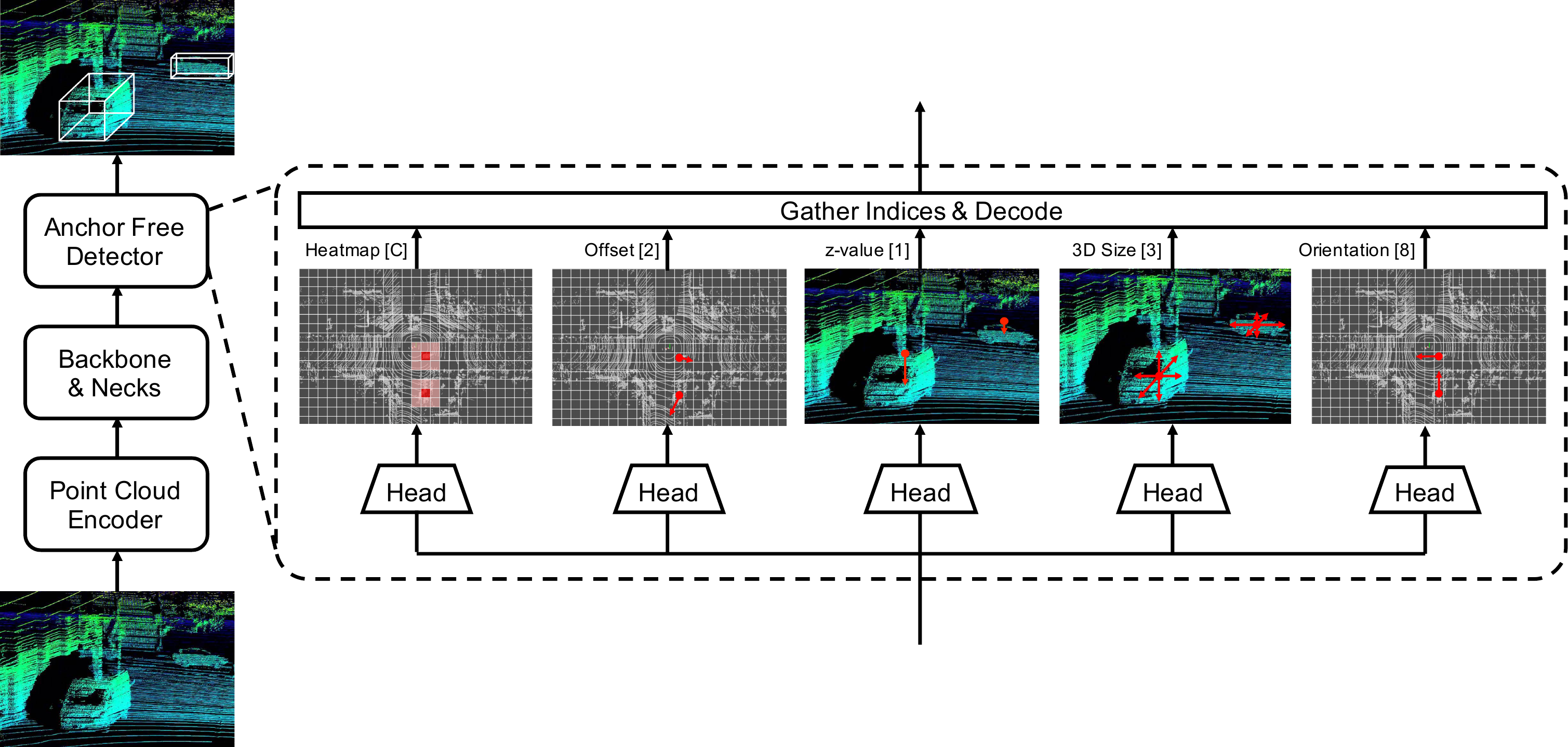}
\end{center}
\caption{The framework of anchor free one stage 3D detection (AFDet) system and detailed structure of anchor free detector. The whole pipeline consists of the point cloud encoder, the backbone and necks, and the anchor free detector. The number in the square brackets indicate the number of last convolution layer's output channels. $C$ is the number of categories used in the detection. Better viewed in color and zoomed in for details.}
\label{fig:framework}
\end{figure*}


\section{Methods}
\label{method}
In this section, we present the details of AFDet from three aspects: point cloud encoder, 
backbone and necks, and anchor free detector. The framework is shown in Figure~\ref{fig:framework}.

\subsection{Point Cloud Encoder}
To further tap the efficiency potential of our anchor free detector, we use PointPillars~\cite{lang2019pointpillars} as the point cloud encoder because of its fast speed. First, the detection range is discretized into pillars in the Bird's Eye View (BEV) plane which is also the $x$-$y$ plane. Different points are assigned to different pillars based on their $x$-$y$ values. Every point would also be augmented to $D = 9$ dimensional at this step. Second, the pre-defined $P$ amount of pillars with enough number of points would be applied with a linear layer and a max operation to create an output tensor of size $F \times P$ where $F$ is the number if output channels of the liner layer in PointNet~\cite{qi2017pointnet}. Since $P$ is the number of selected pillars, they are not one-to-one correspondent with the original pillars in the entire detection range. So the third step is to scatter the selected $P$ pillars to their original location on the detection range. After that we can get a pseudo image $I \in{\mathbb {R}^{W \times H \times F}}$ where $W$ and $H$ indicate the width and height, separately.

Although we use PointPillars~\cite{lang2019pointpillars} as the point cloud encoder, our anchor free detector
is compatible with any point cloud encoders 
which generate pseudo images or image-like 2D data.

\subsection{Anchor Free Detector}
Our anchor free detector consists of five heads. They are keypoint heatmap head, local offset head, $z$-axis location head, 3D object size head and orientation head. Figure ~\ref{fig:framework} shows some details of the anchor free detector.

\textbf{Object localization in BEV.} For heatmap head and offset head, we predict a keypoint heatmap $\mathit{\hat{M} \in \mathbb{R}^{W \times H \times C}}$ and a local offset regression map $\hat{O} \in \mathbb{R}^{W \times H \times 2}$ where $C$ is the number of keypoint types. The keypoint heatmap is used to find where the object center is in BEV. The offset regression map is to help the heatmap to find the more accurate object centers in BEV and also help to recover the discretization error caused by the pillarization process.

For a 3D object $k$ with category $c_k$, we parameterize its 3D ground truth bounding box as 
$\mathit{(x^{(k)}, y^{(k)}, z^{(k)}, w^{(k)}, l^{(k)}, h^{(k)}}, \theta^{(k)})$ 
where $x^{(k)}$, $y^{(k)}$, $\theta^{(k)}$ represent the center location in LiDAR coordinate system, $w^{(k)}$, $l^{(k)}$, $h^{(k)}$ are the width, length and height of the bounding box, and $\theta^{(k)}$ is the yaw rotation around $z$-axis which is perpendicular to the ground. Let $\mathit{[(back, front), (left, right)]}$ denote the detection range in $x$-$y$ plane. To be specific, $\mathit{back}$ and $\mathit{front}$ is along the $x$-axis and $\mathit{left}$ and $\mathit{right}$ is along the $y$-axis in the LiDAR coordinate system. In this work, the pillar in $x$-$y$ plane is always a square. So let $b$ denote the pillar side length. Following~\cite{Law_2018_CornerNet}, for each object center we have the keypoint 
$p = \left( \frac{x^{(k)} - \mathit{back}}{b}, \frac{y^{(k)} - \mathit{left}}{b} \right) \in {\mathbb {R}}^{2}$ 
in BEV pseudo image coordinate. $\tilde{p} = \floor*{p}$ is its equivalent in the keypoint heatmap where $\mathit{\floor*{\cdot}}$ is the floor operation. The 2D bounding box in BEV could be expressed as $\left( \frac{x^{(k)} - \mathit{back}}{b}, \frac{y^{(k)} - \mathit{left}}{b}, \mathit{\frac{w^{(k)}}{b}}, \mathit{\frac{l^{(k)}}{b}}, \theta^{(k)} \right)$. 

For each pixel $\mathit{(x, y)}$ which are covered in the 2D bounding boxes in the pseudo image, we set its value in the heatmap following
\begin{equation}
  M_{x,y,c} =
    \begin{cases}
      1, & \text{if $d = 0$}\\
      0.8, & \text{if $d = 1$}\\
      \frac{1}{d}, & \text{else}
    \end{cases}       
\end{equation}
where $d$ is the Euclidean distance calculated between the bounding box center and the corresponding pixel in the discretized pseudo image coordinates. A prediction $\mathit{\hat{M}_{x,y,c}=}$ $1$ represents the object center and $\mathit{\hat{M}_{x,y,c}=}$ $0$ indicates this pillar is background.

$\tilde{p}$, which represents the object centers in BEV, would be treated as positive samples while all other pillars would be treated as negative samples. Following~\cite{Law_2018_CornerNet, zhou2019objects}, we use the modified focal loss~\cite{lin2017focal}
\begin{equation}
\mathcal{L}_{heat} = -\frac{1}{N}\sum_{x,y,c}
    \begin{cases}
        \left(1 - \hat{M}_{x,y,c}\right)^{\alpha}  \quad\quad\smash{\raisebox{-1.6ex}{\text{if ${M}_{x,y,c} = 1$}}} \\
        \quad\quad \log \left(\hat{M}_{x,y,c}\right),  \\
        \left(1 - M_{x,y,c}\right)^{\beta}  \left(\hat{M}_{x,y,c} \right)^{\alpha} \quad\ \ \smash{\raisebox{-1.6ex}{\text{else}}} \\
        \quad\quad \log \left( 1 - \hat{M}_{x,y,c}\right), \\
    \end{cases}
\end{equation}
to train the heatmap where $N$ is the number of object in the detection range and $\alpha$ and $\beta$ are the hyper parameters. We use $\alpha=2$ and $\beta=4$ in all our experiments.

For the offset regression head, there are two main functions. First, it is used to eliminate the error caused by the pillarization process in which we assign the float object centers to integer pillar locations in BEV as we mentioned above. Second, it plays an important role to refine the heatmap object centers' prediction especially when the heatmap predicts wrong centers. To be specific, once the heatmap predicts a wrong center which is several pixels away from the ground truth center, the offset head has the capability to mitigate and even eliminate several pixels' error to the ground truth object center. 

We select a square area with the radius $r$ around object center pixel in the offset regression map. The farther the distance to the object center is, the larger the offset value becomes.
We train the offset using $L_1$ loss
\begin{equation}
\mathit{\mathcal{L}_{off}} = \frac{1}{N}\sum_{p}\sum^{r}_{\delta=-r}\sum^{r}_{\epsilon=-r} \mathit{\left | \hat{O}_{\tilde{p}} - b \left ( p - \tilde{p} + \left ( \delta, \epsilon\right ) \right ) \right |}
\end{equation}
where the training is only for the square area with side length $2\mathit{r} + 1$ around the keypoint locations $\tilde{p}$. We will discuss more about the offset regression in Section~\ref{experiment}.

\textbf{$\bm{z}$-axis location regression.} After the object localizations in BEV, we only have object $x$-$y$ location. Thus we have the $z$-axis location head to regress the $z$-axis values. We directly regress $z$-value $\hat{Z} \in \mathbb{R}^{W \times H \times 1}$ using $L_1$ loss
\begin{equation}
\mathit{\mathcal{L}_{z}} = \frac{1}{N}\sum_{k=1}^{N} \mathit{\left | \hat{Z}_{p^{(k)}} - z^{(k)} \right |}.
\end{equation}

\textbf{Size regression.} Additionally, we regress the object sizes $\hat{S} \in \mathbb{R}^{W \times H \times 3}$ directly. For each object, we have $s^{(k)} = (w^{(k)}, l^{(k)}, h^{(k)})$. The training loss for size regression is  
\begin{equation}
\mathit{\mathcal{L}_{size}} = \frac{1}{N}\sum_{k=1}^{N} \mathit{\left | \hat{S}_{p^{(k)}} - s^{(k)} \right |}
\end{equation}
which is also the $L_1$ loss.

\textbf{Orientation prediction.} Orientation $\theta^{(k)}$ for object $k$ is the scalar angle rotated around $z$-axis which is perpendicular to the ground. We follow~\cite{Mousavian_2017_CVPR, zhou2019objects} to encode it to an eight scalars with four scalars for each bin. Two scalars are for the softmax classification and the other two are for the angle regression. The angle ranges for two bins are $\Psi_1= \left[ -\frac{7\pi}{6},  \frac{\pi}{6}\right]$ and $\Psi_2 = \left[ -\frac{\pi}{6},  \frac{7\pi}{6}\right]$ which overlap slightly. For each bin, we predict $\hat{\mu}^{(k)}_i \in \mathbb{R}^2$ which are used for softmax classification and $\hat{\nu}^{(k)}_i \in \mathbb{R}^2$ which are used for calculating $\sin$ and $\cos$ value of the offset to the bin center $\gamma_i$. The classification part $\hat{\mu}^{(k)}_i$ is trained with softmax while the offset part $\hat{\nu}^{(k)}_i$ is trained with $L_1$ loss. So the loss for the orientation training is

\begin{equation}
\begin{split}
\mathit{\mathcal{L}_{ori}} = \frac{1}{N}\sum_{k=1}^{N}\sum_{i=1}^{2} \bigg(  \text{softmax}\left(\hat{\mu}^{(k)}_i, \eta^{(k)}_i \right) \\
+ \eta^{(k)}_i \left| \hat{\nu}^{(k)}_i - \nu^{(k)}_i\right| \bigg )
\end{split}
\end{equation}
where $\eta^{(k)}_i = \mathbbm{1} \ (\theta^{(k)} \in \Psi_i)$ in which $\mathbbm{1}$ is the indicator function, and $\nu^{(k)}_i=\left( \sin \left (\theta^{(k)} - \gamma_i \right ),\cos \left(\theta^{(k)}- \gamma_i\right)\right)$. We can decode the predicted orientation value using 
\begin{equation}
\hat{\theta}^{(k)} = \arctantwo \left(\hat{\nu}^{(k)}_{j,1}, \hat{\nu}^{(k)}_{j,2}\right) + \gamma_j 
\end{equation}
where $j$ is the bin index with the larger classification score for object $k$.

\textbf{Loss.} We have described the losses for each head. The overall training objective is
\begin{equation}
\mathit{\mathcal{L} = \mathcal{L}_{heat} + \lambda_{off}\mathcal{L}_{off} + \lambda_{z}\mathcal{L}_{z} + \lambda_{size}\mathcal{L}_{size} + \lambda_{ori}\mathcal{L}_{ori}}
\end{equation}
where $\lambda$ represents the weight for each heads. For all regression heads including local offset, $z$-axis location, size, orientation regression, we only regress $N$ objects which are in the detection range.
\begin{figure}
\begin{center}
\includegraphics[width=0.3\textwidth]{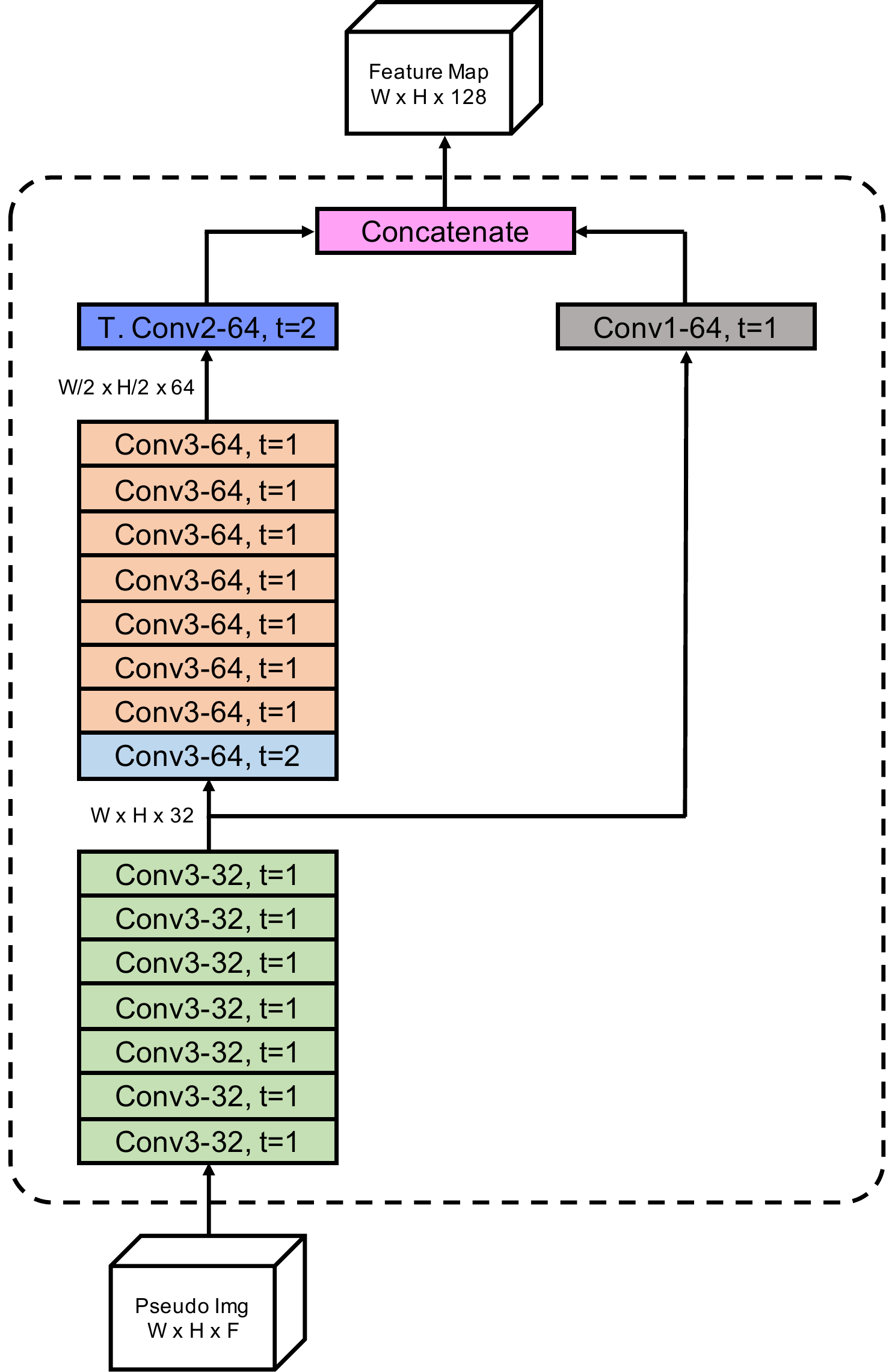}
\end{center}
\caption{The backbone and necks we used in KITTI~\cite{Geiger2012CVPR} detection. Different colors represent different operations with different parameters. The pseudo image is from point cloud encoder. $t$ represents stride. $F$ is the number of channels of the pseudo image. $W$ and $H$ are the width and height, separately. T. Conv is short for transposed convolution. Better viewed in color.}

\label{fig:backbone}
\end{figure}

\textbf{Gather indices and decode.} 
At the training stage, we do not do back-propagation for the entire feature maps. Instead, we only back-propagate the indices that are the object centers for all regression heads. At the inference stage, we use max pooling and AND operation to find the peaks in the predicted heatmap following~\cite{zhou2019objects} which is much faster and more efficient than IoU-based NMS. 

After the max pooling and AND operation, we can easily gather the indices of each center
$\left ( \hat{x}, \hat{y}\right)$ from the keypoint heatmap. Let $\hat{\Phi}$ denote the set of detected BEV object centers. We have
$\hat{\Phi} = \left\{ \left ( \hat{x}^{(k)}, \hat{y}^{(k)}\right)\right\}^n_{k=1}$ where $n$ is the total number of detected objects. Then the final object center in BEV would be $\left(  b \left ( \hat{x}^{(k)} + 0.5 \right) + \hat{o}^{(k)}_1,  b \left ( \hat{y}^{(k)} + 0.5 \right) + \hat{o}^{(k)}_2  \right)$ where $\left (\hat{o}^{(k)}_1, \hat{o}^{(k)}_2 \right)$ are found in the $\hat{O} \in \mathbb{R}^{W \times H \times 2}$ using the index $\left ( \hat{x}^{(k)}, \hat{y}^{(k)}\right)$. For all other prediction values, either they are directly from the regression results or we have mentioned the decoding process above. The predicted bounding box for object $k$ is

\begin{equation}
\begin{split}
\bigg(  b \left( \hat{x}^{(k)} + 0.5 \right) + \hat{o}^{(k)}_1,  b \left ( \hat{y}^{(k)} + 0.5 \right) + \hat{o}^{(k)}_2 , \\
\hat{z}^{(k)}, \hat{w}^{(k)}, \hat{l}^{(k)}, \hat{h}^{(k)}, \hat{\theta}^{(k)} \bigg).
\end{split}
\end{equation}

\subsection{Backbone and Necks}
In this work, we make several key modifications to the backbone used in~\cite{zhou2018voxelnet, yan2018second, lang2019pointpillars} to support our anchor free detector. The network includes the backbone part and the necks part. The backbone part is similar to the network used in the classification tasks~\cite{simonyan2014very} which is used to extract features while downsampling the spatial size through different blocks. The necks part is used to upsample the features to make sure all outputs from different blocks of the backbone have the same spatial size so that we can concatenate them along one axis. Figure~\ref{fig:backbone} shows details of the backbone and necks.

First, we reduce the backbone~\cite{zhou2018voxelnet, yan2018second, lang2019pointpillars} from 3 blocks to 2 blocks. A block $\mathcal{B}(T, E, A)$ consists of $E$ convolution layers with $A$ output channels, each followed by a BatchNorm~\cite{pmlr-v37-ioffe15} and a ReLU. $T$ is defined as the downsampling stride for this block. By reducing the blocks' number from 3 to 2, we remove the feature maps that are downsampled 4 times in~\cite{zhou2018voxelnet, yan2018second, lang2019pointpillars}. We accordingly reduce the upsampling necks $\mathcal{V}(T, A)$ from 3 to 2. Each upsampling neck contains one transposed convolution with $A$ output channels and $T$ upsampling stride followed by BatchNorm and ReLU. Second, the first block we use is $\mathcal{B}(1, 8, 32)$ which does not downsample the output feature size compared with input size.

So the final backbone and necks consists of two blocks $\mathcal{B}(1, 7, 32)$ and $\mathcal{B}(2, 8, 64)$ followed by two upsampling necks $\mathcal{V}(1, 64)$, $\mathcal{V}(2, 64)$, separately. By doing this, the width and height of the input feature maps and the pseudo images are the same. In one word, in the process of generating feature maps we do not downsample, which is critical to maintaining a similar detection performance with ~\cite{lang2019pointpillars} for KITTI~\cite{Geiger2012CVPR} dataset. Reducing downsampling stride will only increase FLOPs, so we also reduce the number of filters in the backbone and necks. It turns out that we have fewer FLOPs in the backbone and necks than~\cite{zhou2018voxelnet, yan2018second, lang2019pointpillars}. We will talk more about the backbone and necks in Section~\ref{experiment}.


\section{Experiments}
\label{experiment}
In this section, we first introduce the two datasets. Then we describe the experiment settings and our data augmentation strategy. Finally, we show the performance on KITTI~\cite{Geiger2012CVPR} validation set and some preliminary results on Waymo~\cite{sun2019scalability} validation set.

\subsection{Datasets}
\textbf{KITTI} object detection dataset~\cite{Geiger2012CVPR} consists of $7,481$ training samples with both calibrations and annotations and $7,518$ test samples which only have calibrations. In our experiments, we split the official $7,481$ training samples into a training set comprising $3,712$ samples and a validation set with the rest $3,769$ samples following~\cite{chen20153d}.
KITTI dataset provides both LiDAR point clouds and images, however, annotations are only labeled in the camera field of view (FOV). To accelerate the training process, we crop out points that are in camera FOV for training and evaluation ~\cite{chen2017multi, zhou2018voxelnet}.

\textbf{Waymo} Open Dataset (Waymo OD)~\cite{sun2019scalability} is a newly released large dataset for autonomous driving. It consists of $798$ training sequences with around $158,361$ samples and $202$ validation sequences with around $40,077$ samples. Unlike KITTI where only the objects in camera FOV are labeled, the objects in Waymo are labeled in the full $360^{\circ}$ field.

\subsection{Experiments Settings}
Unless we explicitly indicate, all parameters showing here are their default values. We use AdamW ~\cite{loshchilov2018decoupled} optimizer with one-cycle policy~\cite{one_cycle}. We set learning rate max to $3 \times 10^{-3}$, division factor to 2, momentum ranges from 0.95 to 0.85, fixed weight decay to 0.01 to achieve convergence. The weight we use for different sub-losses are $\lambda_{off}=1.0$, $\lambda_{z}=1.5$, $\lambda_{size}=0.3$ and $\lambda_{ori}=1.0$. For the following part, we first introduce the parameters used in KITTI~\cite{Geiger2012CVPR}. Then we introduce the Waymo OD parameters that are different from KITTI. 

For KITTI car detection, we set detection range as $\left[ \left( 0, 70.4\right), \left(-40, 40\right), \left(-3, 1\right)\right]$ along $x$, $y$, $z$ axes respectively. So the pseudo images are $I \in \mathbb{R}^{416 \times 480 \times 64}$. This range is the same as PointPillars~\cite{lang2019pointpillars} settings for a fair comparison. We use the max number of objects $50$ which means at most we detect $50$ objects for each class. For PointPillars encoder~\cite{lang2019pointpillars}, we use pillar side length $0.16$ m, max number of points per pillar 100 and max number of pillars $P = 12000$. We set the number of output channels of the linear layer in the encoder to 64. For the backbone, all the convolution layers are with kernel size 3. Their stride and number of output filters are shown in Figure~\ref{fig:backbone}. So the outputs of the backbone and necks are with shape $W \times H \times 128$ which have the same width and height with the pseudo images. For every head, we use two convolution layers: the first convolution layer is with kernel size 3 and channel number 32; the second convolution layer is with kernel size 1. Channel numbers are different for different heads which are shown in Figure~\ref{fig:framework}. For offset regression head, we use $r=2$ as default which means we will regress a square area with side length $5$. We use max pooling with kernel size 3, stride 1 and apply AND operation between the feature map before and after the max pooling to get the peaks of the keypoint heatmaps at the inference stage. So we do not need NMS to suppress overlapped detections. The model is trained for 240 epochs. Due to the small size of the KITTI~\cite{Geiger2012CVPR} dataset, we run every experiment 3 times and select the best one on the validation set.

For Waymo OD vehicle detection, we set detection range as $\left[ \left( -76.8, 76.8\right), \left( -76.8, 76.8\right), \left( -3, 5\right)  \right]$. The max number of objects is set to $200$. The two convolution layers in every head are with channel number 64. For Waymo OD, we use the same backbone as~\cite{zhou2018voxelnet, yan2018second, lang2019pointpillars}.

\subsection{Data Augmentation}
 First, we generate a database containing the labels of all ground truths and their associated point cloud data. For each sample, we randomly select 15 ground truth samples for car/vehicle and place them into the current point cloud. After this, we increase the number of ground truth in one point cloud. Second, each bounding box and the points inside it are rotated following the uniform distribution and translated following the normal distribution. The rotation follows $\mathcal{U}\left ( -\frac{\pi}{20}, \frac{\pi}{20} \right )$ around $z$-axis. The translation follows $\mathcal{N}\left ( 0, 0.25\right)$ for all axes. Third, we also do randomly flip along $z$-axis~\cite{yang2018pixor}, global rotation following $\mathcal{U}\left ( -\frac{\pi}{4}, \frac{\pi}{4} \right )$ and global scaling~\cite{zhou2018voxelnet, yan2018second, lang2019pointpillars}.

\subsection{Evaluation on KITTI Validation Set}
We follow the official KITTI evaluation protocol to evaluate our detector, where the IoU threshold is 0.7 for the car class. We compare different methods or variants using average precision ($\mathit{AP}$) metric.

We first compare different heatmap prediction methods and different offset regression methods. Then we compare the different backbone for our detector. Finally, we compare our method with PointPillars.

\begin{figure}
\begin{center}
\includegraphics[width=0.47\textwidth]{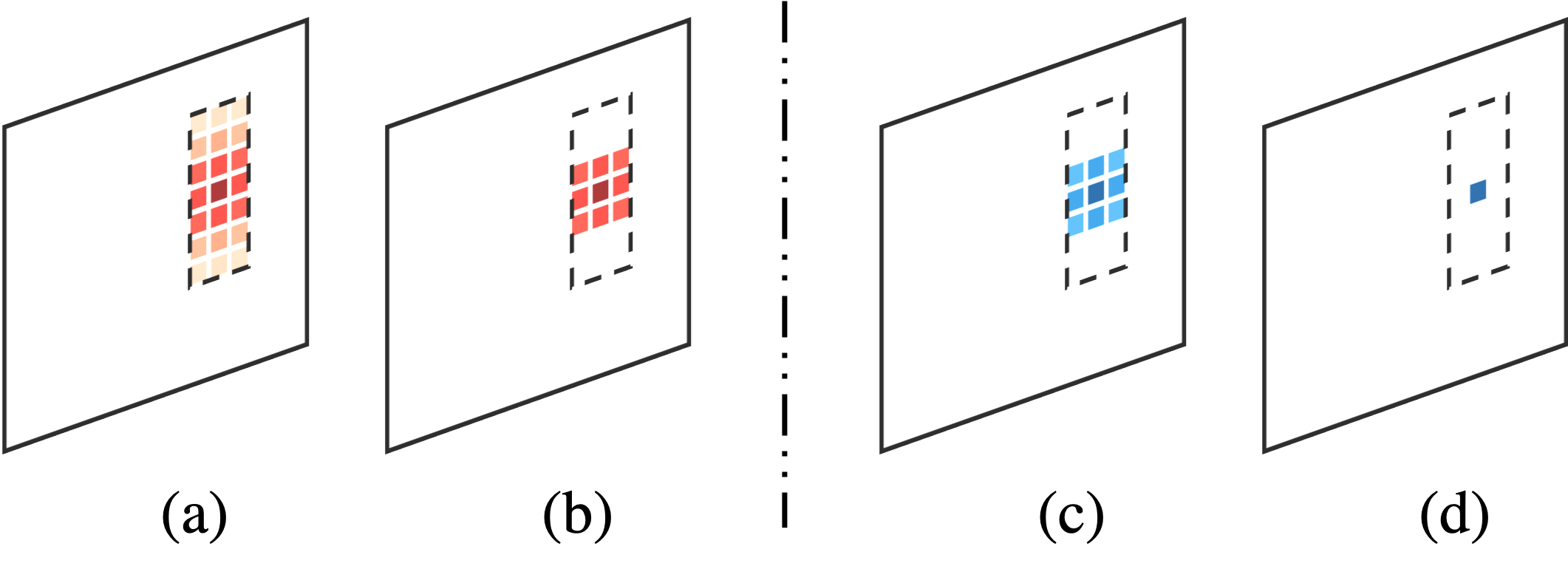}
\end{center}
\caption{(a) is the car shape heatmap prediction method and (b) is the Gaussian kernel heatmap prediction method. (c) is our offset regression method and (d) is the regression method in~\cite{zhou2019objects}. The left two rectangles represent the heatmap outputs and the right two rectangles represent the offset regression outputs. The dashed line rectangles indicate the 2D bounding boxes.}

\label{fig:car_shape}
\end{figure}

\begin{table}[t]
\begin{center}
\begin{tabular}{l|c c c}
\hline
 \multirow{2}{*}{Methods} &\multicolumn{3}{|c}{3D $\mathit{AP}$ IoU=0.7} \\
  &Mod &Easy &Hard \\
\hline
\hline
Gaussian Kernel &72.50 &82.57 &68.91   \\
Car Shape (Ours) &\textbf{75.57} &\textbf{85.68} &\textbf{69.31}   \\
\hline
\hline
$r=0$~\cite{zhou2019objects} &74.51 &84.45 &69.03   \\
$r=1$ &74.76 &85.16 &68.84   \\
$r=2$ &\textbf{75.57} &\textbf{85.68} &\textbf{69.31}   \\
$r=3$ &73.63 &78.80 &68.53   \\
\hline
\end{tabular}

\end{center}
\caption{The comparison between the two heatmap prediction methods and the comparison for different regression area radius.}
\label{tbl:heatmap_comparison}
\end{table}



\begin{figure*}
\center

\includegraphics[width=0.9\textwidth]{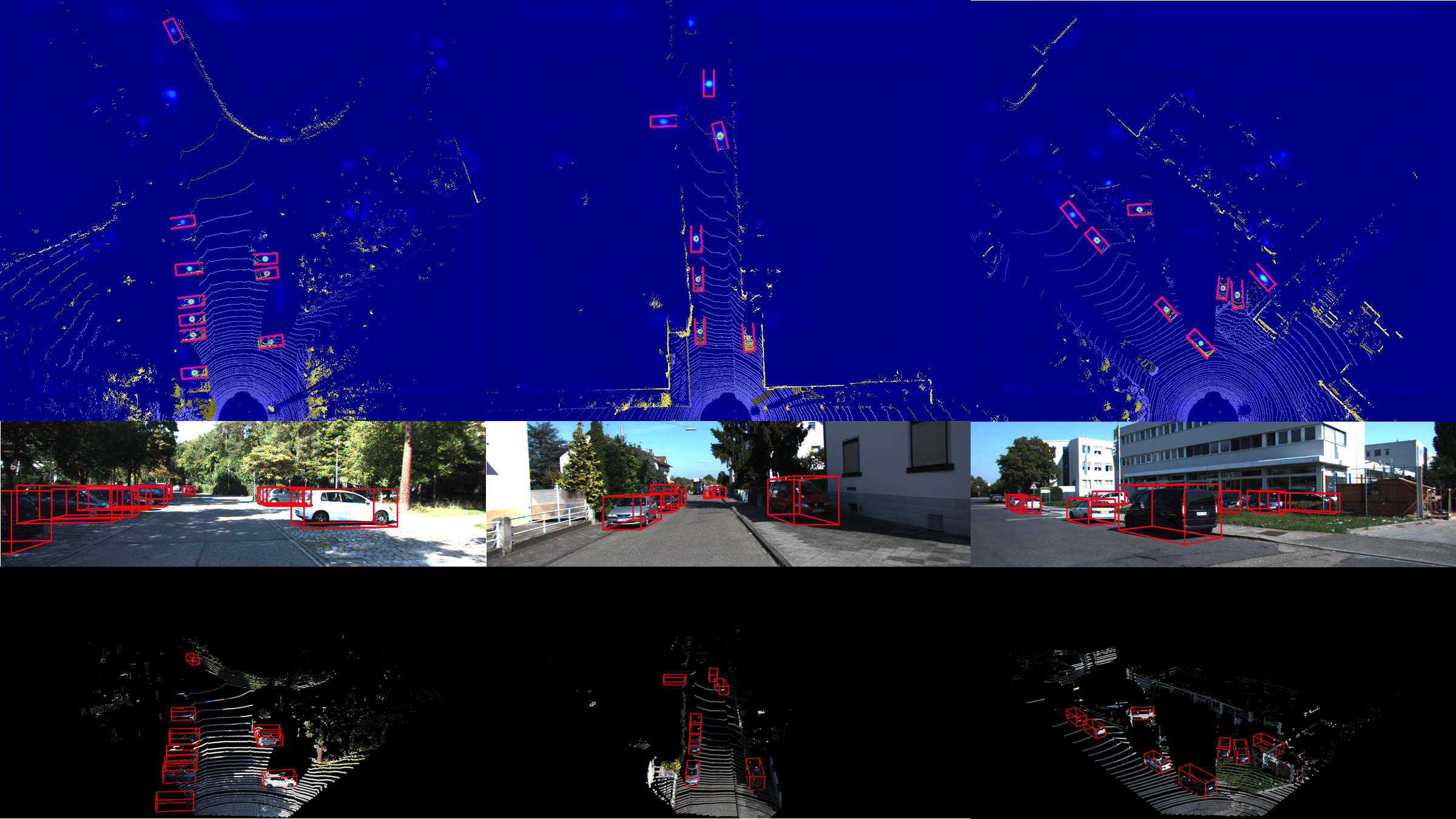}
\caption{Results visualization on KITTI car detection with AFDet. Each one consists of heatmap, projection results in 2D RGB image and 3D point cloud results from top to bottom. Better viewed in color and zoomed in for details.}
\label{fig:vis}
\end{figure*}

\begin{table*}[t]
\begin{center}
\vspace{5pt}
\setlength\tabcolsep{4pt}
\begin{tabular}{l|c|c|c|c c c|c c c}
\hline
 \multirow{2}{*}{Methods} &\multirow{2}{*}{\# Params} &\multirow{2}{*}{\# MACs} &\multicolumn{1}{|c|}{Anchor} &\multicolumn{3}{|c|}{3D $\mathit{AP}$ IoU=0.7} &\multicolumn{3}{|c}{BEV $\mathit{AP}$ IoU=0.7}\\
  & & &Free &Mod &Easy &Hard &Mod &Easy &Hard\\
\hline
\hline
PointPillars~\cite{lang2019pointpillars}          &4.81$\mathit{M}$ &62.22$\mathit{G}$ &\xmark &76.04 &83.73 &69.12 &86.34 &89.68 &84.38   \\
\hline
$\mathcal{B}(2, 4, 64) + \mathcal{B}(2, 6, 128)  + \mathcal{B}(2, 6, 256) $           &5.91$\mathit{M}$ &125.37$\mathit{G}$ &\cmark  &72.62     &81.01     &67.47     &82.72 &87.10 &78.97   \\
$\mathcal{B}(1, 4, 64) + \mathcal{B}(2, 6, 128)  + \mathcal{B}(2, 6, 256) $           &5.91$\mathit{M}$ &501.46$\mathit{G}$ &\cmark &75.33     &85.18     &69.18     &84.69 &88.91 &79.83   \\
$\mathcal{B}(1, 7, 32) + \mathcal{B}(2, 8, 64) $           &0.56$\mathit{M}$  &76.53$\mathit{G}$ &\cmark &75.57     &85.68     &69.31     &85.45 &89.42 &80.56   \\

\hline
\end{tabular}
\end{center}
\caption{The KITTI~\cite{Geiger2012CVPR} validation set car detection performance comparison between different variants of AFDet and reimplemented PointPillars. The \# parameters and \# MACs are calculated on the entire network including backbone and necks and detector but except for the point cloud encoder. The \# parameters and \# MACs in the point cloud encoder are same for all listed methods above.} 
\label{tbl:performance}
\end{table*}

\textbf{Heatmap prediction.} We compare our car shape heatmap prediction method with the Gaussian heatmap prediction method~\cite{zhou2019objects}. For the car shape heatmap prediction, we have described in Section~\ref{method}. For the Gaussian heatmap prediction, we splat all ground truth keypoints onto a heatmap $\mathit{M \in \mathbb{R}^{W \times H \times C}}$ using a Gaussian kernel $\mathit{M_{x,y,c}}=\exp \left({-\frac{(x-\tilde{p}_{x})^2 + (y-\tilde{p}_{y})^2}{2\sigma^{2}_{p}}}\right)$ where ${\mathit{\sigma_p}}$ is the size adaptive standard deviation from~\cite{Law_2018_CornerNet}. The biggest difference between the two methods is the number of non-zero predictions in the heatmap. For the Gaussian kernel method, the non-zero predictions are only several pixels (\eg 9 pixels) around the object center in the heatmap. While for the car shape method, all pixels in the 2D bounding box (car shape in BEV view) are non-zero. The illustration could be found in Figure~\ref{fig:car_shape} (a) and (b). From Table~\ref{tbl:heatmap_comparison}, we can see that predicting the entire car shape rather than the Gaussian kernel can improve about 2\% on moderate difficulty.

\textbf{Offset regression.} To verify the effectiveness of our proposed offset regression method in which the training is for the square area with side length $2\mathit{r}+1$ around the object center $\tilde{p}$, we compare it with the offset regression method proposed in~\cite{zhou2019objects} in which the training is only for the object center $\tilde{p}$. Actually the latter regression method~\cite{zhou2019objects} is a special case of our method when $r$ equals 0. The illustration of two methods is shown in~\ref{fig:car_shape} (c) and (d). We set $r$ to 0, 1, 2 and 3. From Table~\ref{tbl:heatmap_comparison}, we can see that by setting $r$ to 2. We can achieve 1 $\mathit{AP}$ improvement over the regression method mentioned in~\cite{zhou2019objects}.

\begin{table*}[t]
\begin{center}
\begin{tabular}{l|c|c|c c c c}
\hline
\multirow{2}{*}{Methods} &\multicolumn{1}{|c|}{Anchor} &\multirow{2}{*}{\# Epochs}  &\multicolumn{4}{|c}{LEVEL\_1 3D $\mathit{AP}$ IoU=0.7}\\
 &Free & &Overall &0 - 30m &30 - 50m &50m - $\infty$\\
\hline
\hline

StarNet~\cite{ngiam2019starnet}    &\xmark &75 &53.70 &- &- &-  \\
PointPillars\footnotemark~\cite{lang2019pointpillars}      &\xmark &100 &56.62 &81.01 &51.75 &27.94 \\
PPBA~\cite{cheng2020improving}+PointPillars  &\xmark &- &62.44 &- &- &- \\
MVF~\cite{zhou2019end}        &\xmark &100 &62.93 &86.30 &60.02 &\textbf{36.02} \\
\hline
AFDet+PointPillars-0.16 (Ours)  &\cmark  &16 &58.77 &84.99 &55.76 &24.78 \\
AFDet+PointPillars-0.10 (Ours)  &\cmark &16 &\textbf{63.69} &\textbf{87.38} &\textbf{62.19} &29.27 \\
\hline
\end{tabular}

\end{center}
\caption{The vehicle detection performance comparison for single-stage 3D detection methods on Waymo OD validation set.} 
\label{tbl:waymo_performance}
\end{table*}

\textbf{Backbone and necks.} We made some modifications in the backbone and necks for KITTI~\cite{Geiger2012CVPR}. The baseline of our method is termed  as $\mathcal{B}(2, 4, 64) + \mathcal{B}(2, 6, 128)  + \mathcal{B}(2, 6, 256) $  in Table~\ref{tbl:performance} which is same to~\cite{lang2019pointpillars}.

First, the backbone used in~\cite{lang2019pointpillars} is downsampled 3$\times$ with stride 2 for each block. After upsampling, the feature map size used in the detection head is downsampled by 2$\times$ compared with the pseudo images. We remove the first downsampling stride and keep the following downsampling stride which is shown as $\mathcal{B}(1, 4, 64) + \mathcal{B}(2, 6, 128)  + \mathcal{B}(2, 6, 256) $ in Table~\ref{tbl:performance}. The feature map sizes to the detector are the same as the pseudo images. We can see that the performance improves around 2\% compared with the baseline. But the \# MACs improve from 125.37$\mathit{G}$ to 501.46$\mathit{G}$ which is about $4\times$ of calculation of the baseline. This is mainly caused by doubling the feature maps' width and height.

Second, by modifying downsampling stride the performance improves. But we need to make sure that the performance improvement comes from enlarging the feature map size rather than from increasing computation. So we reduce the number of downsampling blocks from 3 to 2, in which we remove the last downsampling block. We also halve the number of output filters in the convolution layers. This computation reducing modification is shown as $\mathcal{B}(1, 7, 32) + \mathcal{B}(2, 8, 64) $ in Table~\ref{tbl:performance}. We can see that the performance has nearly no change by reducing the computation. From  $\mathcal{B}(1, 4, 64) + \mathcal{B}(2, 6, 128)  + \mathcal{B}(2, 6, 256) $ to $\mathcal{B}(1, 7, 32) + \mathcal{B}(2, 8, 64) $, we reduce about $84\%$ \# MACs and about $90\%$ \# parameters. So enlarging the feature map in our anchor free detector helps to improve the performance.

\footnotetext{\cite{zhou2019end, ngiam2019starnet, cheng2020improving} report slightly different performance on the same method. Here we adopt the results reported in~\cite{zhou2019end}.}

\textbf{Comparison with PointPillars.} We compare our method with PointPillars~\cite{lang2019pointpillars} on KITTI validation set. We use Det3D\footnotemark~\cite{zhu_det3d} implementation to evaluate PointPillars~\cite{lang2019pointpillars}. All comparisons are under the same settings including but not limited to detection range and PointPillars size. As we can see, our AFDet with the modified backbone $\mathcal{B}(1, 7, 32) + \mathcal{B}(2, 8, 64) $ can achieve similar performance with PointPillars~\cite{lang2019pointpillars}. But our method does not have a complex post-processing process. We do not need the traditional NMS to filter out results. More importantly, the \# parameters in AFDet is about 0.56$\mathit{M}$, which is only about 11.6$\%$ of its equivalent in PointPillars~\cite{lang2019pointpillars}. 

\footnotetext{\url{https://github.com/poodarchu/Det3D}}

Furthermore, using max pooling and AND operation rather than NMS would make it more friendly to deploy AFDet on the embedded systems. We can run nearly the entire algorithm on a CNN accelerator without the tedious post-processing on CPU. We could reserve more CPU computation resources for other tasks in autonomous driving cars. We also tried kernel sizes 5 and 7 in the max pooling. It does not show much difference with kernel size 3.

We show three qualitative results in Figure~\ref{fig:vis}. As we can see, AFDet has the capability to detect the object centers in the heatmap. It can also regress other object attributes (\eg object sizes, $z$-axis locations and others) well. We validate the effectiveness of the anchor free method on 3D point cloud detection.

\subsection{Preliminary Results on Waymo Validation Set}

We also include some preliminary evaluation results on Waymo OD~\cite{sun2019scalability} validation set. We use Waymo online system to evaluate our performance.  We try our best to have the same settings and parameters for a fair comparison. But sometimes we do not know other methods' detailed parameters. On Waymo OD, we train our model with significantly less number of epochs compared with other methods. But we still show competitive or even better results.

We show two AFDet results with PoinPillars~\cite{lang2019pointpillars} encoders in Table~\ref{tbl:waymo_performance}. The number after the encoder name represents the voxel size in $x$-$y$ plane. As we can see, our ``AFDet+PointPillars-0.16'' with voxel size 0.16 m beats ``PointPillars'' by 2\% on LEVEL\_1 vehicle detection. When we reduce the voxel size to 0.10 m, our ``AFDet+PointPillars-0.10'' outperforms the state-of-the-art single-stage methods on Waymo validation set. We only train our model for 16 epochs while others train their models for 75 or 100 epochs for better convergence.

\section{Conclusion}

In this paper, we tried to address the 3D point cloud detection problem. We presented a novel anchor free one stage 3D object detector (AFDet) to detect the 3D object in the point cloud. We are the first to use anchor free and NMS free method in 3D point cloud detection which has the advantage in the embedded systems. All experimental results proved the effectiveness of our proposed method.


{\small
\bibliographystyle{ieee_fullname}
\bibliography{egbib}
}

\end{document}